# An optical biomimetic "eyes" with interested object imaging

Jun Li[1,*,✉], Shimei Chen[1,*], Shangyuan Wang[1], Miao Lei[1], Xiaofang Dai[1], Chuangxue Liang[1], Kunyuan Xu[2], Shuxin Lin[2], Yuhui Li[1], Yuer Fan[1], Ting Zhong[1]

We presented an optical system to perform imaging interested objects in complex scenes, like the creature easy see the interested prey in the hunt for complex environments. It utilized Deep-learning network to learn the interested objects's vision features and designed the corresponding "imaging matrices", furthermore the learned matrixes act as the measurement matrix to complete compressive imaging with a single-pixel camera, finally we can using the compressed image data to only image the interested objects without the rest objects and backgrounds of the scenes with the previous Deep-learning network. Our results demonstrate that no matter interested object is single feature or rich details, the interference can be successfully filtered out and this idea can be applied in some common applications that effectively improve the performance. This bio-inspired optical system can act as the creature eye to achieve success on interested-based object imaging, object detection, object recognition and object tracking, etc.

Object imaging, Single-pixel camera, Deep-learning network, Compressive imaging

Visual information is usually one of the most important information in the interaction of the human being and the world. As the alternative of the human eye, visual imaging system widely used for machine vision[1], video imaging[2], robot vision[3], autonomous vehicles[4], etc. Since the first camera come to being by Daguerre on 1839, all kinds of imaging system sprang up. Additionally, Kevin F. Kelly of Rice University et al. proposed the single-pixel camera (SPC) to achieve high-resolution imaging, which greatly reduces the amount of data collection in the imaging process[5]. Since then, SPC has been used in various imaging applications[6,7]. Throughout these visual imaging systems, we first acquired the whole scene information, then detected, recognized, tracked interested objects. That is to say, there is many image data to processing to acquire the user's interested object information, the result of high computational expense, poor real-time performance, and great amount data.

But when we discover and amazing the creature's eyes can easy see and track the prey in the hunt for complex environments, without the disturbance of the rest of other objects and backgrounds[8,9], we naturally hope its instinct mechanism and function can be introduced into physics imaging system for overcoming the above problem. As early as 1959, D.H. Hubel first recognized the receptive fields of single neurones in the cat's striate cortex[10-12]. In addition, Z.Kourtzi discussed the representation of perceived object shape by the human lateral occipital complex[13], these research all discover attention mechanism underlying the intrinsic vision mechanism of the creature. Recently Le Chang et al. tried to use primate facial stimulation to reveal the facial recognition codes of the brain, and verified the interpretability of machine codes from the architecture of biological motivation[14]. Based on these works, many researchers introduce this vision attention mechanism to the applications such as medicine[15], computer vision[16], natural language processing[17].

[1]Guangdong Provincial Key Laboratory of Quantum Engineering and Quantum Materials, School of Physics and Telecommunication Engineering, South China Normal University, Guangzhou, China. [2]School of Physics and Telecommunication Engineering, South China Normal University, Guangzhou, China. *These authors contributed equally to this work. ✉e-mail: lijunc@126.com

On the other hands, the people always try to discover the simple principles of object perception[18, 19]. More recently, there are some enlightening works addressing the decoding of seen and imagined objects using hierarchical visual features[20, 21]. Early studies used signal processing to extract primary image features based on intuitive perceived features of human vision. Color histograms[22], color moments[23], and color correlogram[24] are currently the most widely used color features of the objects. Canny et al.[25] proposed the edge detection method of Canny Edge detector to represent the contour feature of the objects. Texture, as a local feature that describes the repeatability and regularity features of the objects, including texture descriptors such as local binary patterns[26], gray-level co-occurrence matrix[27] and fractal functions[28], is also one of important features of the imaging object. D.H BALLARD proposed the generalized Hough transform which simple shape mapping allows the construction of complex shapes[29]. Related findings indicate that the abstraction of visual representations formed along the ventral stream is constantly increasing[30-33]. Also, convolutional neural networks (CNNs) have a powerful ability of learning shallow features and abstract features[35-37, 47]. Most of all adopt CNNs to processing the signal data for simulating the process mechanism of the attention mechanism of thecreature[12, 34]. With CNNs presented, it widely used to learn the object feature information in various fields[35-37].

Here we report a general method combining deep learning to achieve feature extraction of objects of interest, then "learned" features use as structured patterns for illumination in a single-pixel camera to achieve the screening of specific target objects, namely interested object imaging. In particular, instead of studying common image segmentation problems, our goal here is to take a new approach to solve the traditional full-scene imaging problems. It exploits the fact that visual selective attention is equivalent to separating two-dimensional information about the target object from a multi-object scene. So only one part of interested object should be left in the label image. This will result in the range of neural network learning being limited to the target object. The proposed method present this long-term natural-evolving visual attention mechanism in signal form to achieve the effect of simulated biological visual systems on information processing and it can provide solutions for the challenges faced by some vision-based applications. Meanwhile, it breaks through the limitations of traditional imaging methods of image retro-reconstruction, which may provide a new idea for research in the imaging field.

# The proposal of object imaging

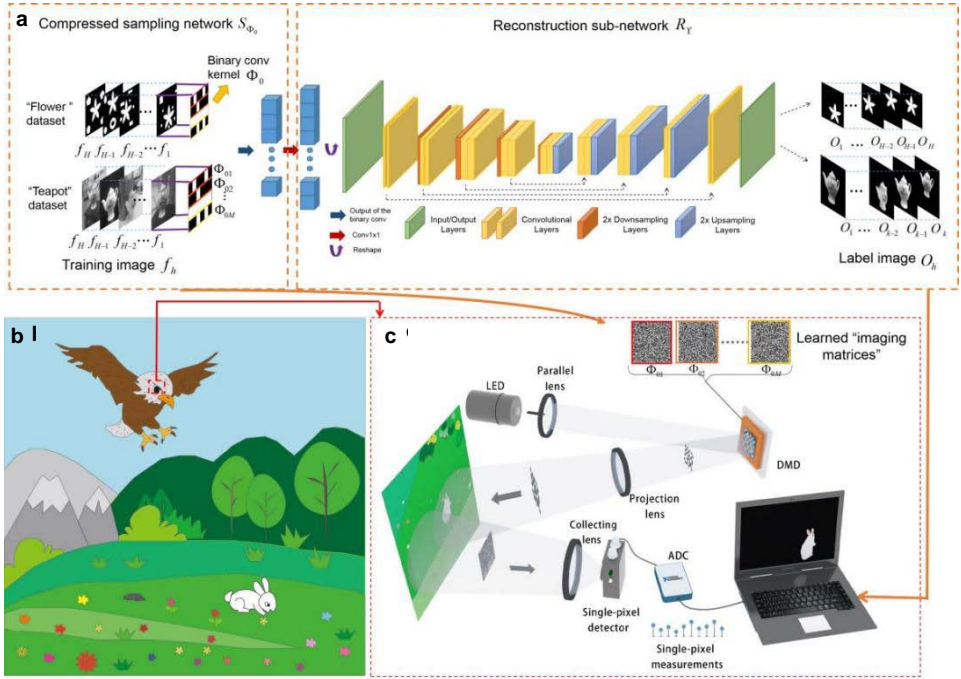

**Fig.1|System design for object imaging which inspired by the visual attention mechanism. a,** Object imaging network. Conv, convolution. **b,** An example of the visual attention mechanism: the eagle hunts for the rabbit. The design system based on this mechanism is implemented in **c**. **c,** Biotic object imaging system based on single-pixel camera. The learned imaging matrices is obtained by **a**.

The mapping relationship between the input and output of the dataset and the difference of the network model will lead to the completely different functions realized in the end. More importantly, the establishment of the dataset plays a key role in the function realization of object imaging. Specifically, according to functional magnetic resonance imaging, it is found that attended objects in biological vision are encoded with high precision, while the ignored objects have no measurable coding[9]. Inspired by this study of biological vision, the label images are designed to mark the objects of interest，and other interfering objects or background are filtered out.

To facilitate the neural network for the specific task, and taking into account the differences between different samples, we select simple binary image — Flower, and the more complex and scene variable grayscale image — Teapot as inputs, which are respectively single features and contain rich details. Teapot is the real shot dataset and Flower is virtual dataset added to different noise interference (see Methods for dataset details). Through the innovative processing of removing irrelevant information to get the label images from the input images, our approach no longer need to learn from all objects like the traditional training methods, but learns only the entire image features contained by the target object, finally extracts the one or many interested object according to fused multi-feature information from the complex scenes (see feature extraction of method).

Our network is the combination of compressed sampling module and U-net structure for automatic feature information extraction and reconstructing the object of interest (Fig.1a). While the network is completely trained by our above specified output label images of the datasets, the compressed sampling network of our network will produce the appropriate structured nose-liked patterns that can effectively acquire the vision feature information of the interested objects, which can be called "imaging matrices". Then, these "imaging matrices" are binarized to adapt to the hardware requirements for optical imaging such as micro-mirror devices (DMD) (Fig.1c). As such, the desired object information is projected into one-dimensional(1-D) data using our designed measurement matrices in biotic object imaging system based on SPC. The detected 1-D signals containing only the feature information of the target object are fed to the trained subsequent reconstruction sub-network for accurate reconstruction. The entire system does not reconstruct the entire image, but only to image specific interested object.

## Neural network of object imaging

Our model uses convolutional neural networks to extract the prior information of target objects in complex scenes. The network is mainly composed of two sub-networks, ie, compressed sampling network and reconstruction sub-network. In detail, the sampling network is trained to obtain binary "imaging matrices" containing only the feature of target object, and the reconstruction sub-network is used to reconstruct the target object image from the 1-D measurement signal.

Formally, the convolution process in the deep learning can be expressed as follow:

$$y = \sigma(W * f + b), \qquad (1)$$

where $f \in R^{N \times N}$ stands for the input image, $W \in R^{N \times N}$ represents the convolution kernel or weight matrix, $b$ is the bias, $\sigma$ is the activation function, $y$ is the feature map after convolution, and $*$ indicates the convolution operation.

By linear projection, compressive sensing(CS) can capture a compressible or sparse signal into a compressed form：

$$y = \langle \Phi, f \rangle, \qquad (2)$$

where $\Phi \in \mathbb{R}^{M \times N} (M \leq N)$ is the measurement matrix, and $\langle \cdot \rangle$ denote the inner product. Without activation function and bias, the convolution operation can be regarded as the linear projection of CS[46]. Thus, deep learning, an end-to-end learning method, can be used to train to obtain measurement matrix for specific tasks.

The compressed sampling network adaptively learns the "imaging matrices" according to the train dataset, which is actually a binary convolutional layer.[2] Moreover, the weight value of the convolution kernels is limited to $\{-1, +1\}$ in the training stage. When the network is optimized, we can obtain the learned binary

"imaging matrices". Indeed, it can effectively extract the interested object features in the scene, leading the measurement signal retains more object information, so as to better reconstruct the target object and filter out the background information and the interference of non-target objects. In biotic object imaging system based on SPC, "imaging matrices" are used as structured patterns for illumination to complete the biotic visual object imaging system.

The reconstruction sub-network is non-linear deep network, which is used to recover 2-D target object image from 1-D measurement signal. The measurement signal $y$ is firstly convolved into feature map of $N \times N \times 1$ by convolutional layer with convolution kernel of 1x1, and then is reshaped into feature map of $N \times N$, and finally through U-net depth network to recover 2-D interested target object image.

In the training step, the compressed sampling network and the reconstruction sub-network form an end-to-end network for joint optimization. This joint optimization can be learned after training the network from $H$ pairs of different label training data, and each pair has a known only specific object label image $O_h$ and training image $f_h$ of this object in complex scene, where $h = 1, 2, ..., H$. The training process is similar to the optimization process and can be expressed as:

$$\{\Phi_0, \Upsilon\} = \arg\min_{\Phi_0, \Upsilon} \frac{1}{H} \sum_{h=1}^{H} \| O_h - \tilde{O}_h \| \tag{3}$$

$$\tilde{O}_h = R_\Upsilon(S_{\Phi_0}(f_h)) \tag{4}$$

where $S_{\Phi_0}$ is the compressed sampling network, $R_\Upsilon$ is the nonlinear reconstruction sub-network, and $\Phi_0$ and $\Upsilon$ denote the weight of the compressed sampling network and the weight of the reconstruction sub-network, respectively. $\tilde{O}_h$ represents reconstructed object image using optimized network model in the prediction process $\|\cdot\|$ is a loss function about the error between $O_h$ and $\tilde{O}_h$. We set the learning rate to 0.0002 and use stochastic gradient descent (SGD) and Adam optimizer to optimize and update parameters of model.

All programs are running under Pytorch in Python3.7 environment, and accelerated calculations are performed using NVIDIA Geforce GTX1080Ti GPU. The trained model can well filter out simple and complex interference of simple objects (Flower), and can still be perfectly reconstructed even when the orientation and rotation angle be changed (Fig.2a). For complex objects (Teapot) in the case of changing in size, placement, shooting angle, and light intensity, it can effectively remove simple interference, complex interference, artificial scenes, and background interference in natural scenes of complex objects (Teapot) (Fig.2b). Naturally, the feature of Flower is simple and easy to train, so both PSNR and SSIM values are higher than those of Teapot (Fig.2c,d). We have also performed noise robustness analysis on Teapot, and when multiple Teapot are changed in size, or flipped, our method can still image almost perfectly (Extended Data Fig.1a,b,c).

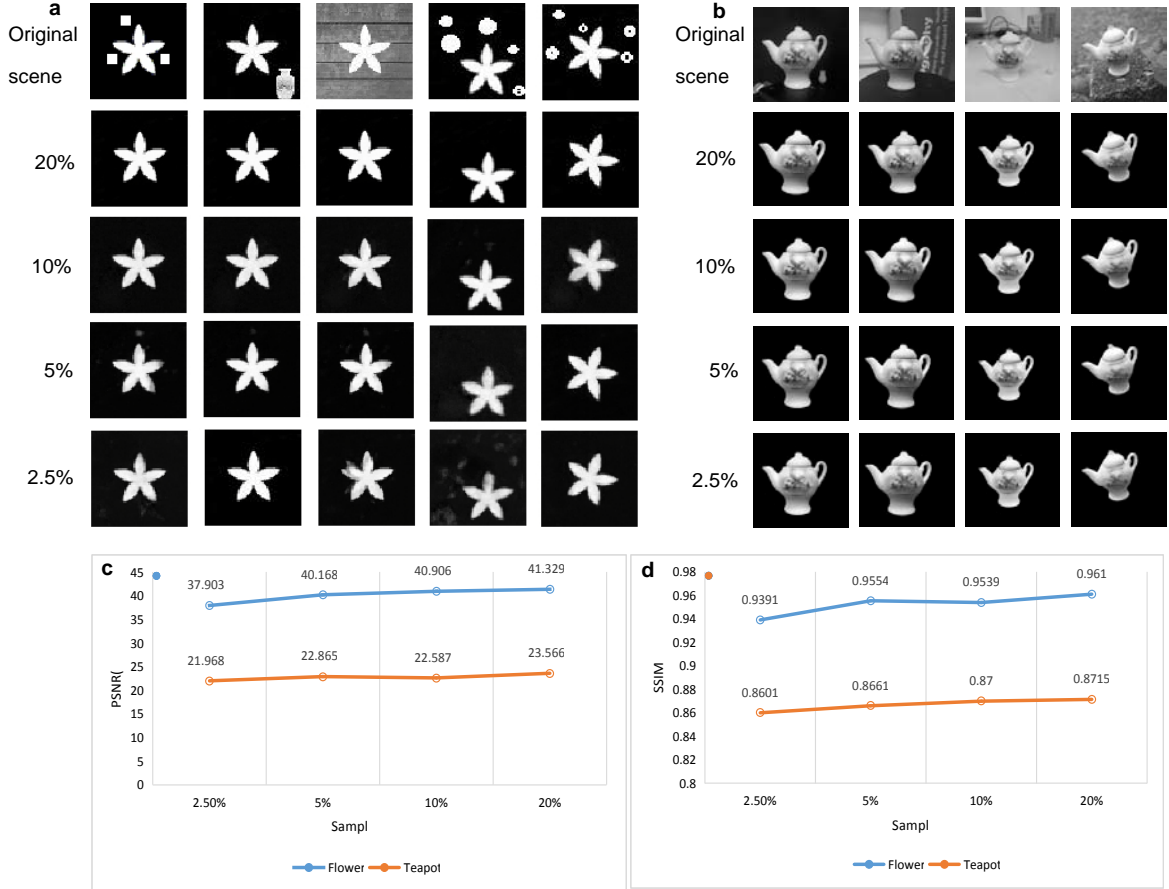

**Fig.2|Simulation results and performance evaluation of the target object of interest. a,** Simulation results of Flower at different sampling rates. From left to right are the results of simple interference without background, slightly complex vase interference without background, complex background, simple interference after translation without background, and simple interference after rotation without background. **b,** Simulation results of Teapot at different sampling rates. From left to right are the results of simple interference, complex interference, artificial background, natural background. **c,** The average PSNR values between the reconstructed images and the label image under different sampling rates. PSNR, peak signal-to-noise ratio. **d,** The average SSIM values between the reconstructed images and the label image under different sampling rates. SSIM, structural similarity index measure.

## Experimental system of object imaging

The experimental setup is shown in Fig.1c. We use the "Flower" as the target object. A 3-watt white LED as the light source, modulates the scattered light into parallel light through a parallel lens and then illuminates DMD (Texas Instruments discovery 4100). A series of binary "imaging matrices" trained by CNN are loaded into the DMD. The DMD displays different patterns in sequence. To make full use of the DMD area of 1024x768, the 64x64 pixel "imaging matrices" have been expanded tenfold. With these structured patterns illuminate the scene in turn, we use a photodiode (Thorlabs PDA36A) as single-pixel detector to collect the light intensity signal. Then, the light intensity signal is converted into digital signal by the data acquisition panel (National Instruments PCI-4474) and fed to the optimized reconstruction sub-network on the computer for only reconstructing target object. We only performed experiments on Flower (Fig.3a). We can see that all interferences are filtered out, and the final imaging result is only Flower, which initially verifies the feasibility of our proposed method.

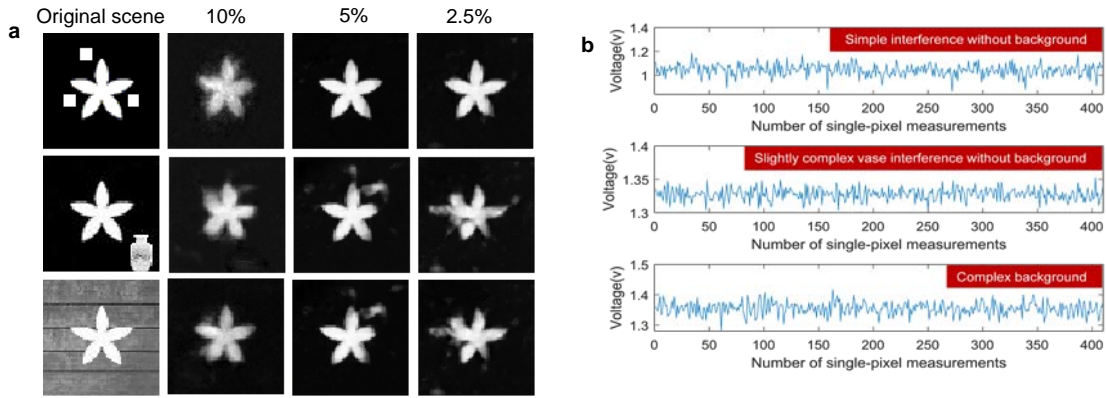

**Fig.3|The experiment results of Flower and the corresponding experimental data. a,** The experimental results of Flower at different sampling rates. From top to bottom are the results of simple interference without background, slightly complex vase interference without background and complex background. **b,** The original data detected by single-pixel detector at 10%.

## Examples for applying object imaging

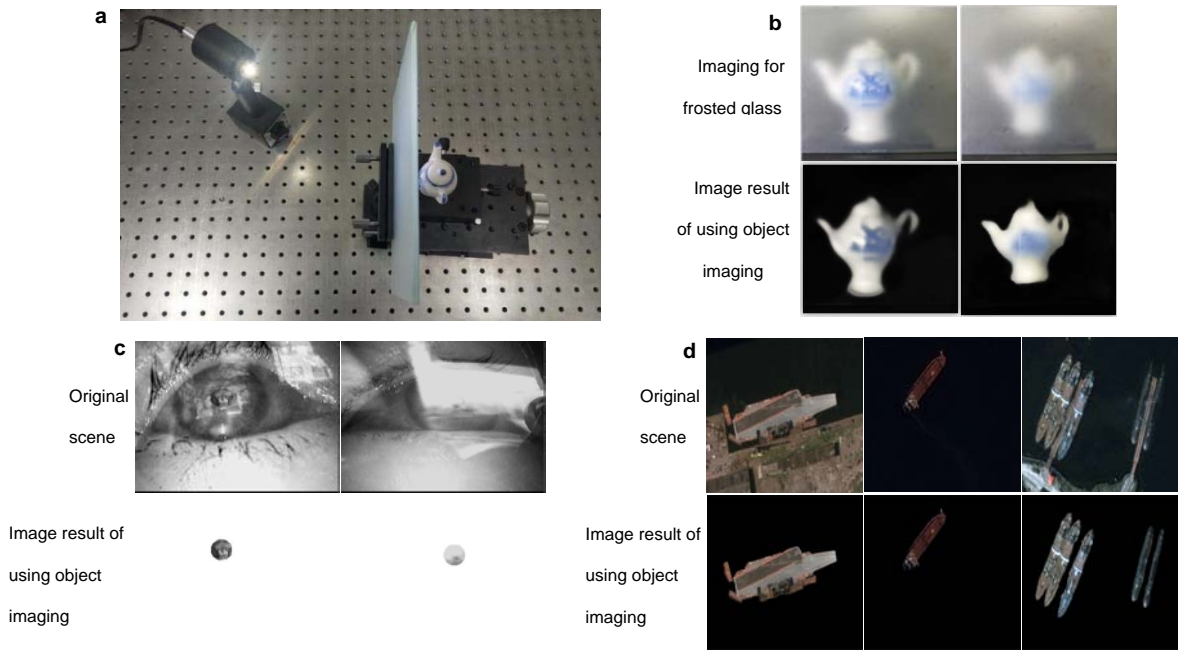

**Fig.4|Object imaging is applied in different examples. a,** Imaging through scattering media. **b,** Imaging results of the scattering medium. From left to right are the original image and the corresponding imaging result. **c,** The imaging result for pupil positioning using object imaging. **d,** The imaging result for vessel target detection using object imaging.

**Table 1|Comparison of different methods of pupil positioning**

|  | ElSe | PupilNet | DeepEye | Our |
|---|---|---|---|---|
| Average Detection Rate(%) | 69.27 | 78.63 | 86.83 | 89.93 |

**Table 2|Performance comparison with yolov3 in vessel target detection**

|  | YOLO3 | Our |
|---|---|---|
| Mean Average Precision(%) | 63.89 | 73.55 |
| Accuracy(%) | 52.30 | 62.60 |
| Recognition rate（fps） | 2.1697 | 2.5961 |

The proposed method based on imitating the mechanism of visual attention can be applied in the following challenging scenarios and is implemented under pure numbers.

In practice, it is difficult to identify the approximate outlines of objects in opaque scattering medium[48]. Due to the low visibility, the imaging quality is greatly affected, and even imaging is impossible. Our method can be applied here. Frosted glass as scattering medium, is placed in the target object (Teapot) to simulate the scattering media scene (Fig.4a). As a result, the target object resolution in the captured image is low and even hard to identify its general profile. The Teapot dataset remains unchanged. Here, the results show that the proposed method can both overcome the difficult of imaging and realize to filter out the optical base in the scattering medium (Fig.4b).

Eye tracking is an important technology for obtaining the user's gaze direction, and is widely used in psychology[38], human-computer interaction[39] and other research areas[40]. Then, pupil positioning is a pivotal link in gaze tracking, so its effectiveness and accuracy deeply affect the functional realization of the whole human-machine interaction system. We use 24 datasets[41, 42], a total of 70,000 pictures, take the pupil as specific object, retain the area inside the circle, and filter out the interference area outside the circle. The Euclidean distance is less than 5 as the basis for correct pupil detection (Fig.4c). Comparing the final calculation result with Else algorithm[43], DeepEye[44], PupilNet[45], it can be found that the pupil detection rate has been improved (Table 1).

Another application is target detection in remote sensing images of ships. Vessel target detection is of great significance to sea cruises[49]. Using our method, we can solve the problem of consuming time and energy caused by the use of human eyes to identify ships. Under the same conditions, with 800 samples(Fig.4d), we also tested the classic method—yolov3[50], and analyzed and compared the performance of the two methods from the three methods of mean Average Precision(mAP), accuracy, and recognition rate. From the results in the Table 2, compared with the traditional algorithm model, our method increases mAP by 9.66%, accuracy by 10.30%, and speed by 0.4264fps.

## Discussion

Our approach simulates the visual attention of biological vision for active screening of interested targets. The proposed method effectively extracts the characteristic information of the object of interest to filter out interference and actively screen the object of interest. However, due to the network training in the theoretical noise-free environment, the system noise will lead to the deviation of the "imaging matrices" in the practical experimental system. In order to reduce the impact of the noise, in the next work, we will consider introducing the experimental data of real optical imaging system into the training. As such, it not only further improves the image quality but also allows for imaging with complex objects. Compared with traditional target recognition methods, this method not only breaks through the limitation of imaging first and then recognition processing, but also reduces the analysis and processing of a large amount of useless information. The practicability of this method will have important and far-reaching application prospects in all traditional imaging fields and high-resolution video applications can be implemented in the future (Supplementary Videos 1, 2). In future work, we will consider implementing this kind of bionic vision system on hardware to make a bionic eye camera.

# Methods

In the natural scene $f$, when the eagle catches the rabbit (Fig.1b), it will see M objects, assuming that each object is represented by $f_i (i=1,...M)$. Then, for M objects, the whole scene, $f$ is $\sum_{i=1}^{M} f_i$.

Assuming that there are N image features which are taken from primary to abstraction. Different features are independent of each other and obey the Gaussian distribution. After the normalization, they are mapped into feature vectors:

$$\begin{cases} (e_{11}, e_{12}, ..., e_{1n}) = \xi_1 \\ \vdots \\ (e_{k1}, e_{k2}, ..., e_{kn}) = \xi_K, \\ \vdots \\ (e_{n1}, e_{n2}, ..., e_{nn}) = \xi_N \end{cases} \qquad (5)$$

where $(e_{11}, e_{12}, ..., e_{1n})$ is the feature vector mapped by primary feature; as image feature are increasingly complex, $(e_{n1}, e_{n2}, ..., e_{nn})$ is the feature vector mapped by abstract feature.

Then the N-dimensional feature space X is formed by the vector $\xi_1, \cdots \xi_K, \cdots \xi_N$

$$X = span\{\xi_1, \cdots \xi_K, \cdots \xi_N\} \qquad (6)$$

Certainly, every object $f_i$ can be represented by this set of vectors.

$$f_i = \sum_{j=1}^{N} \xi_j x_j, \qquad (7)$$

where $x_1, x_2, ..., x_N$ are the feature decomposition coefficient. Thus, a scene image can be expressed as:

$$f = \sum_{i=1}^{M} f_i = \sum_{i=1}^{M} \sum_{j=1}^{N} \xi_{ij} x_{ij}. \qquad (8)$$

However, among the many objects, the eagle's target of interest is only the rabbit $f_0$, so the feature of the rabbit can be identified and extracted to accurately capture it.

In order to obtain only interested objects $f_0$, the neural network is used to learn the feature of interested objects, and the weight of compressed sampling network is optimized (equation(3)) to get the imaging matrix of the target object and the detailed form of the $\Phi_0$:

$$\Phi_0 = \sum_{n=1}^{M} \omega_n \cdot \begin{bmatrix} \xi_1 \\ \vdots \\ \xi_K \\ \vdots \\ \xi_N \end{bmatrix}, \qquad (9)$$

where $\omega_1, \omega_2, \cdots, \omega_M$ is the specific coefficient which is the learned weight parameters and this imaging matrix $\Phi_0 \in \mathbb{R}^{M \times N} (M \leq N)$. Essentially, according to the different feature of the learned object, the specific object matched by $\Phi_0$ is different. Finally, the scene image is illuminated by the designed imaging matrices, and the specific object can be obtained according to the extracted feature information.

$$\Phi_0 f = \begin{bmatrix} \sum_{n=1}^{M} \omega_n \xi_1 \\ \vdots \\ \sum_{n=1}^{M} \omega_n \xi_K \\ \vdots \\ \sum_{n=1}^{M} \omega_n \xi_N \end{bmatrix} \cdot \sum_{i=1}^{M} \sum_{j=1}^{N} \xi_{ij} x_{ij} = \sum_{j=1}^{N} \xi_{0j} x_{0j} = \hat{f}_0 \quad (10)$$

Finally, biotic object imaging system based on SPC is used to achieve this process. The "learned" object features extracted through network is displayed on the DMD as the structured patterns, and the detected 1-D data that contains feature information for the object of interest is fed back to the reconstruction sub-network to reconstruct the final interested targets.

**Details on dataset acquisition**

In order to better verify the effectiveness and universality of the proposed method, this paper fully considers the differences of different samples. In the process of creating the data set, two different data sets are selected for verification.

Flower--Flower is from the "device" image set from the MPEG7_CE-Shape-1_part_B sample library. We used it as the base image to add simple distractions such as squares, triangles, and other basic patterns, and shifted them up, left, right, and rotated them by 20 degrees to create a virtual dataset of 1000 with 64x64 pixels (Extended Data Fig.2a). This is used to illustrate that our method has the advantages of translational, rotation invariant in extracting features.

Teapot--We chose Teapot with rich details and obvious contours as samples. Place the sample on a black circular rotating table, and use a fixed angle iPhone 6s to take a shot in gray mode every 5 degrees. A total of 72 training samples are obtained, which are expanded to 87 by adding interference backgrounds (Extended Data Fig.2c). In addition, we use vivo X20A and put the Teapot in the life scene for shooting, and the selected scenes are randomly and diverse, from simple to complex, from single to multiple. In order to be able to simulate the effectiveness of the method in this article on the target object in all scenes, the scenes we photographed are more extensive and have a certain degree of randomness, such as grass and trees in natural scenes, lakes, and man-made scenes such as office desktops, equipment accessories in optical laboratories, houses, etc. At the same time, the shooting angle is also random to simulate different states of specific objects, such as different positions, different sizes, and different shapes (Extended Data Fig.2b). A total of 210 images were collected during the shooting process. The sample images were all color RGB images with a size of 2000×2000 pixels. The dataset was expanded to 2160 through data enhancement operations such as translation and motion blur. In order to meet the requirements of the optical experimental platform, they are all converted into single-channel grayscale images, and each training sample image is compressed into 128x128 pixels through bicubic interpolation.

**Comparison of different methods**

Without setting the compression rate, our method is compared with the reconstruction results using the machine learning PCA, ICA method in previous work.(Extended Data Fig.3a) Since the selected target object is a real object, both the main feature and the local feature are more complex and abstract, which also requires the selection of more effective feature extraction techniques. The results show that the three methods can still extract the feature information of the specific object Teapot, but the reconstruction effect of the PCA and ICA methods is very poor. This is because PCA and ICA are only linear expressions of feature information, and the extraction of more complex object features is relatively limitations, unable to extract higher-dimensional information of the image. The jump connection structure used in the U-shaped network model integrates the shallow main body information and the deep local information, and feature extraction is more effective. Thus, it can extract and recognize the target object under the complex background, and the resolution effect is very ideal, which is far better than PCA and ICA methods(Extended Data Fig.3b).

**Analysis on the learned "imaging matrices"**

In practice, it is difficult to verify whether a certain measurement matrix satisfies RIP, so the coherence theory is used to weaken RIP to analyze the measurement matrix. The mutual coherence $\mu(\Psi)$ represents the worst case coherence between any two columns of and is one of the most fundamental quantities associated with CS theory. For a measurement matrix $\Psi \in R^{M \times N}$, the mutual coherence can be defined as:

$$\mu(\Psi) = \max_{1 \leq i \leq j \leq N} |\langle \phi_i, \phi_j \rangle|, \qquad (11)$$

where $\phi_i$ is the $i$ column of $\Phi$ and $\mu(\Psi) \in \left( \sqrt{\frac{N-M}{M(N-1)}}, 1 \right)$. The mutual coherence of the measurement matrix of the two objects at different compression rates is shown in the following table. From the results, the designed matrix is satisfied with the CS theory. (Extended Data Table 1)

**Acknowledgements** This work was supported by the Project of Natural Science Foundation of Guangdong Province, China (No. 2015A030313384), Science and Technology Program of Guangzhou, China (No. 201607010275).

**Author contributions** J.L. conceived the ideas of the work and the proposed framework. S.C., S.W., M.L., and X.D. contributed to design dataset. S.C. worked on optical experiments with Flower. S.C. performed the example of scattering media and the comparison of different methods. K.X. and S.L.

performed the example of pupil positioning. Y.F. and Y.L. realized the example of vessel target detection. C.L. was involved in the design of the proposed network. M.L. and T.Z. contributed to verify the matrix. J.L. and S.C. led the writing and revision of the manuscript. All authors discussed ideas and results, and contributed to the manuscript.

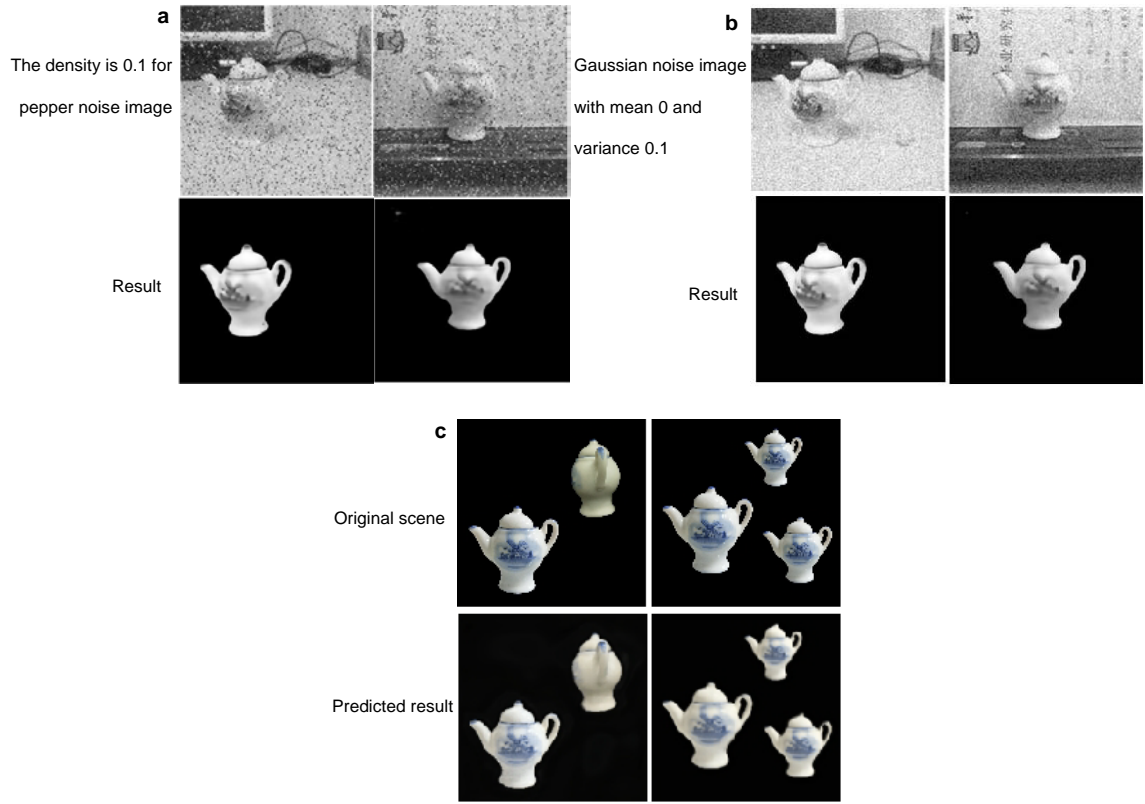

**Extended Date Fig.1|The noise robustness results of Teapot and the imaging of multiple interested objects. a,** Imaging results under pepper noise. **b,** Imaging results under gaussian noise. **c,** Imaging results in multi-target scenary.

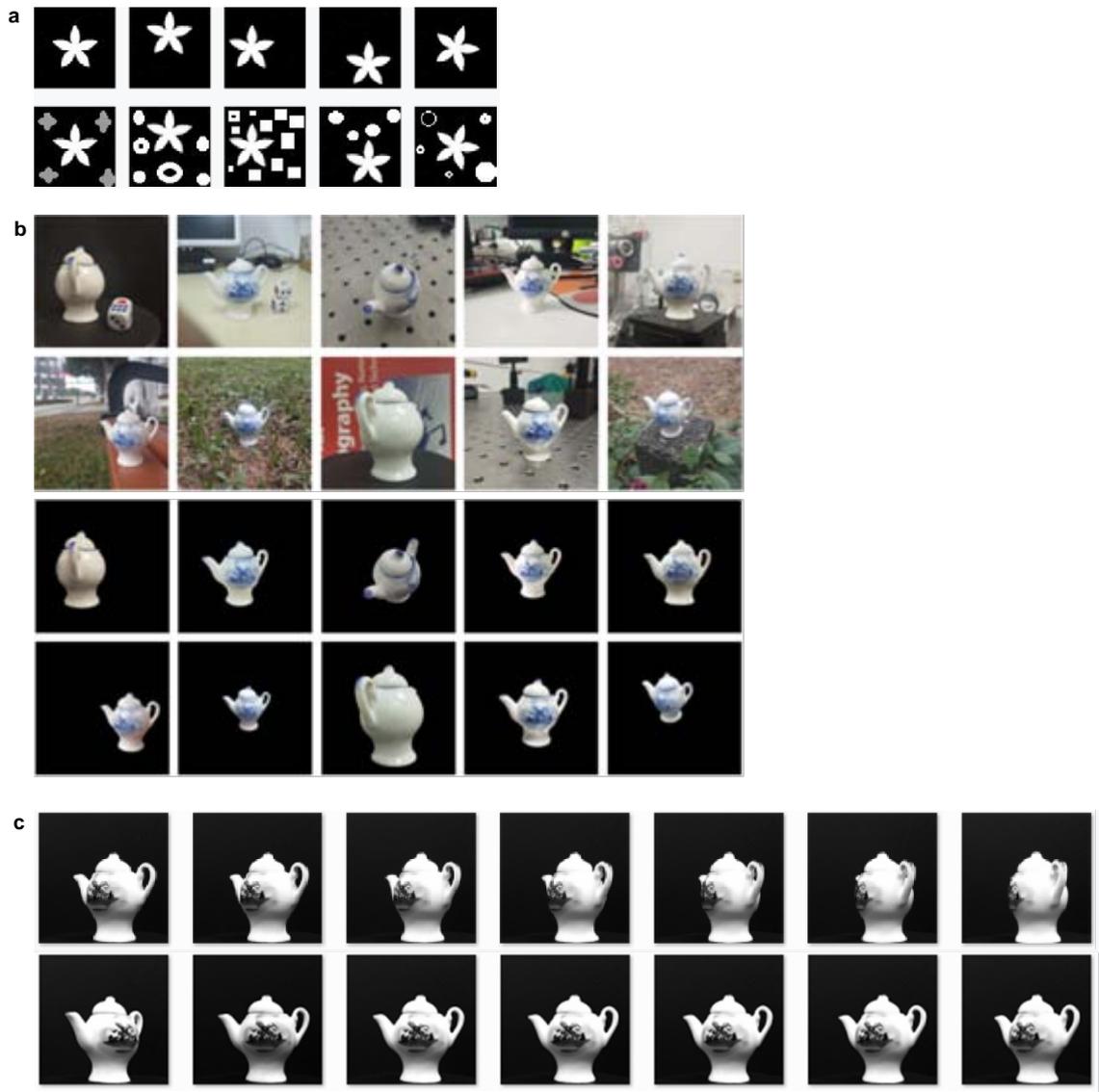

**Extended Date Fig.2|Details of the dataset. a,** Design details of Flower dataset. **b,** The dataset of Teapot in real scene shot. **c,** The dataset of Teapot in rotation.

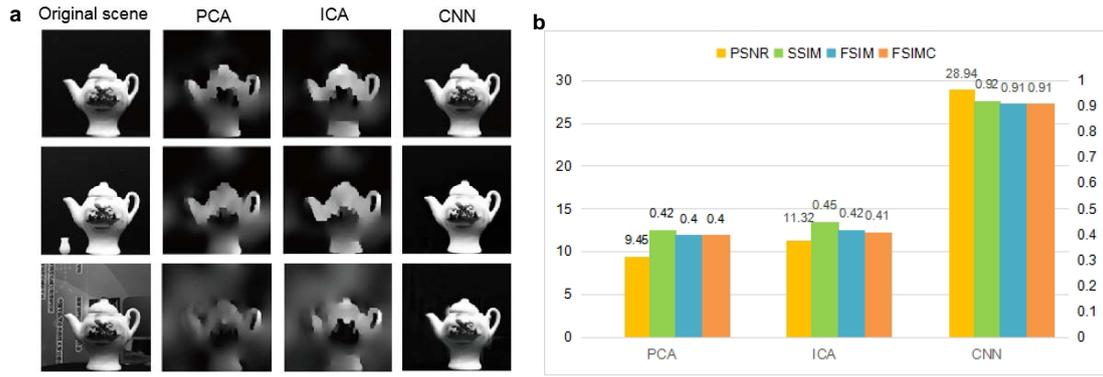

**Extended Date Fig.3|Performance evaluation of different methods. a**, Comparison of the experimental results of the three methods of PCA, ICA and CNN. **b**, Image quality evaluation. PSNR, peak signal-to-noise ratio. SSIM, structural similarity index measure. FSIM, feature similarity index measure. FSIMC, feature similarity of color.

**Extended Data Table 1 | The mutual coherence at different sampling rates**

| Sampling rate | 0.2 | 0.1 | 0.05 | 0.025 |
|---|---|---|---|---|
| Flower | 0.1852 | 0.2581 | 0.3861 | 0.4952 |
| Teapot | 0.1004 | 0.1413 | 0.2086 | 0.2976 |